\documentclass{article}

\PassOptionsToPackage{numbers, compress}{natbib}

\usepackage[preprint]{neurips_2020}

\usepackage[utf8]{inputenc} 
\usepackage[T1]{fontenc}    
\usepackage{hyperref}       
\usepackage{url}            
\usepackage{booktabs}       
\usepackage{amsfonts}       
\usepackage{nicefrac}       
\usepackage{microtype}      

\usepackage{enumerate}
\usepackage{graphicx}
\usepackage{wrapfig}

\title{Attention or memory? \\ Neurointerpretable agents in space and time}

\author{
  Lennart Bramlage\\
  Faculty of Technology\\
  Bielefeld University\\
  \texttt{lbramlage@techfak.de} \\
  \And
  Aurelio Cortese \\
  Computational Neuroscience Labs \\
  ATR Institute International, Japan \\
  \texttt{cortese\_a@atr.jp} \\
}

\usepackage{float}
\usepackage{caption}
\usepackage{subcaption}

\begin{document}

\maketitle

\begin{abstract}
In neuroscience, attention has been shown to bidirectionally interact with reinforcement learning (RL) processes. This interaction is thought to support dimensionality reduction of task representations, restricting computations to relevant features. However, it remains unclear whether these properties can translate into real algorithmic advantages for artificial agents, especially in dynamic environments. We design a model incorporating a self-attention mechanism that implements task-state representations in semantic feature-space, and test it on a battery of Atari games. To evaluate the agent’s selective properties, we add a large volume of task-irrelevant features to observations. In line with neuroscience predictions, self-attention leads to increased robustness to noise compared to benchmark models. Strikingly, this self-attention mechanism is general enough, such that it can be naturally extended to implement a transient working-memory, able to solve a partially observable maze task. Lastly, we highlight the predictive quality of attended stimuli. Because we use semantic observations, we can uncover not only which features the agent elects to base decisions on, but also how it chooses to compile more complex, relational features from simpler ones. These results formally illustrate the benefits of attention in deep RL and provide evidence for the interpretability of self-attention mechanisms.
\end{abstract}

\section{Introduction}
Attention is an oft-discussed and highly popular subject in both recent neuroscience and machine learning literature. While specifics elude both fields to this day, in its essence, attention is the directed focus on task-relevant aspects of the environment while ignoring the rest \citep{Gazzaley2012-qj}. The ability to attend task-relevant features holds many benefits. Notably, animals and humans do not only learn new tasks quickly from noisy stimuli and experiences, but they adapt to changing environments with unmatched dexterity. As recently hypothesized, this could (partly) arise from abstractions and selective attention creating lower-dimensional task-state representations \citep{Niv2019-rb, Cortese2019-hv}. That is, representing a task in terms of only its relevant stimulus dimensions.
In machine learning literature, attention models have been effectively applied to a vast number of problems. Visual selective attention has enabled sequential methods for image processing, similar to saccades in the human eye \citep{Xu2015-nu, Mnih2014-pb}. This approach does not only provide lower-dimensional observations but allows practitioners to interpret the model's process of "seeing.". On the other hand, self-attention \citep{Bahdanau2014-qs} has revolutionized natural language processing by all but replacing more complex recurrent models \citep{Vaswani2017-ns}. Although disputed \citep{Jain2019-we}, this application lends a sense of interpretability to the decision-making process of deep neural networks \citep{Wiegreffe2019-fl, Zambaldi2018-xt, Chaudhari2019-zd}. However, only recently have explicit models of attention broken into the field of (reinforcement) learning for decision-making in dynamic environments. These past applications tend to implement visual, rather than multimodal feature-based attention, generally because they are compared to well-known RL solutions for Atari benchmarks. Yet, recent research in neuroscience suggests that attention mechanisms in the human brain also select a variety of features, stemming not only from sensory inputs, but also from memory, knowledge and predictive forward-modeling \citep{Martinez-Trujillo2004-cj, Farashahi2017-ux, Chun2011-ae, Chun2011-ji}. To investigate current state-of-the-art ML models of attention through the lens of contemporary neuroscience theories, we thus apply the popular self-attention mechanism \citep{Bahdanau2014-qs, Vaswani2017-ns} to a feature-based RL task suite and evaluate its performance in settings with additional noisy features. In all settings, self-attention succeeds in selecting task-relevant over irrelevant features. We observe that while self-attention may not always reach the convergence speed of simpler baseline models, it demonstrates increased resilience towards noisy task-settings.
Furthermore, there is increasing evidence that working memory and attention are grounded in similar mechanisms \citep{DeBettencourt2019-tp, LaBar1999-bv}, and it has been suggested that the two constructs, in fact, largely overlap \citep{Gazzaley2012-qj, Chun2011-ae, Chun2011-ji}. In light of these presumptions, we frame the self-attention mechanism, which has shown great promise in sequence modeling, as a working memory module in RL - whereby attention focus should be directed in time. While our working-memory agent can solve a partially observable T-maze task inspired by rodents’ experiments, we find that it demonstrates the theorized property by maintaining an internal focus on task-relevant past observations. Finally, we show that patterns of self-attention scores illustrate how features and their relations influence behavior.
The main contributions of this paper are:
\begin{enumerate}[I]
    \item we implement a selective attention mechanism in semantic feature space, coupled with reinforcement learning,
    \item we show that the proposed model achieves high resiliency to noise, as predicted by neuroscience empirical and theoretical work
    \item we blur the line between attention and memory by directly extending the attention mechanism to focus on past observations, as a working memory module,
    \item we frame the ubiquitous self-attention mechanism in the broad context of neuroscientific theories of attention.

\end{enumerate}

\section{Related Work}

Our work draws on multidisciplinary ideas from both neuroscience and machine learning.

\textit{Neuroscience} -  The idea that attention and RL are tightly linked goes back to the early studies on classical conditioning in animals, in the 1970s and 1980s \citep{Mackintosh1975-mz, Pearce1980-gh}. Later work has extensively investigated how rewards or task goals teach attentional search \citep{Sepulveda2020-lm, Chelazzi2013-qq, Anderson2017-im}, and promote attention-driven plasticity in sensory cortices \citep{David2012-la, Tankelevitch2020-lu, Arsenault2013-zy}. In parallel, studies in humans and animals have shown how RL depends on attention when task settings become more complex \citep{Niv2015-od, Oemisch2019-wt}. These two lines of evidence have recently been consolidated into a single principle: neuroscientists have demonstrated that in the human brain attention and RL operate in a two-way interaction \citep{Niv2015-od, Leong2017-ps}. In these studies, human participants had to solve problems in multidimensional environments. Learning models fitted to participants’ choices, as well as model-based analysis of neuroimaging data, find that humans pay attention to relevant features of the environment - ignoring the rest, but also that this attention weighting is refined over time through experience. Attention chooses what to learn about, while (reward-based) learned information guides attention. \citet{Farashahi2017-ux} further show that humans learn the value of individual features in dynamic, multidimensional environments, so as to reduce the dimensionality but also to increase behavioural adaptability, particularly in cases when feature values are changing over time.

\textit{Machine Learning} - There is a rich literature on attention models in machine learning. Here we will make explicit note of models of attention in domains for decision making in dynamic environments. Among the earliest examples is the Deep Attention Recurrent Q-Network \citep{Sorokin2015-dw}, which represents CNN-extracted feature patches across channels as weighted sums of all other patches. Importance weights are generated by a two-layer neural network conditioned on both the current observation, and the hidden state of a subsequent recurrent neural network, which is used to generate Q-values. The same method of preattentive feature extraction is used by \citet{Zambaldi2018-xt} and can be understood as encoding an image region in terms of the features it contains. Zambaldi et al. used a self-attention mechanism to relate spatially disparate image regions to each other, thus alleviating the spatial bias of the pre-attentive image encoder. The "what"-"where"-dichotomy is handled differently in Mott et al. \citep{Mott2019-xs}, where a fixed spatial basis is both appended and attended in a separate channel of the image stimulus. Unlike prior work, Mott et al. further suggest an explicit disparity of bottom-up and top-down attention by querying key-representations of an input image with a recurrent network, whose only inputs are prior attention-weighted embeddings of the input. It is, however, difficult to fully support this argument, since the entire architecture is trained end-to-end using actor-critic and thus optimizing an approximate value-function. Arguably, the top-down queries are as much motivated by the current task as the bottom-up image representations. In a rare counterexample, \citet{Yuezhang2018-rw} propose a multiplicative attention mechanism, grounded in optical flow, which is generally understood to be a factor of bottom-up saliency in human vision \citep{Yan2018-no, Rauss2013-xq}. Beyond the sensory, attention has found wide application in models of differentiable memory \citep{Graves2014-am, Graves2016-ha}. Graves et al. use attention mechanisms to read from and write to external memory banks; however, these complex, differentiable computers followed the simple assumption that memory is, at its core, attention over time \citep{Graves2013-ss}. Manchin et al. state that intermittent attention layers in a convolutional network implement rudimentary memory \citep{Manchin2019-un}, because flickering game objects are still tracked through attention. Nevertheless, this is more likely to be due to the network learning these dynamics and an effect observable in agents that do not implement attention explicitly.

We consider a soft self-attention mechanism in feature space, closely following the task-state representation framework \citep{Niv2019-rb}. Our assumption, based on biological precedence, is that the brain represents environmental features, regardless of the learner's task-proficiency. Therefore, we feed the agent a large representation-volume as observations, incorporating features and their expressions across multiple modalities, which may or may not be task-relevant. We expect the model to select task-relevant features for further relational analysis, as theorized by Zambaldi et al., and create compound features that are then fed directly into value- and policy-function \citep{Zambaldi2018-xt}. While we skip the feature-extraction step, our model can be described as a flipped relational inductive bias, allowing for reasoning beyond vision. Following the framework of the task-state representation, we consider features and their expressions as distinct features, requiring different representations.

\begin{figure}
    \centering
    \includegraphics[width=0.8\linewidth]{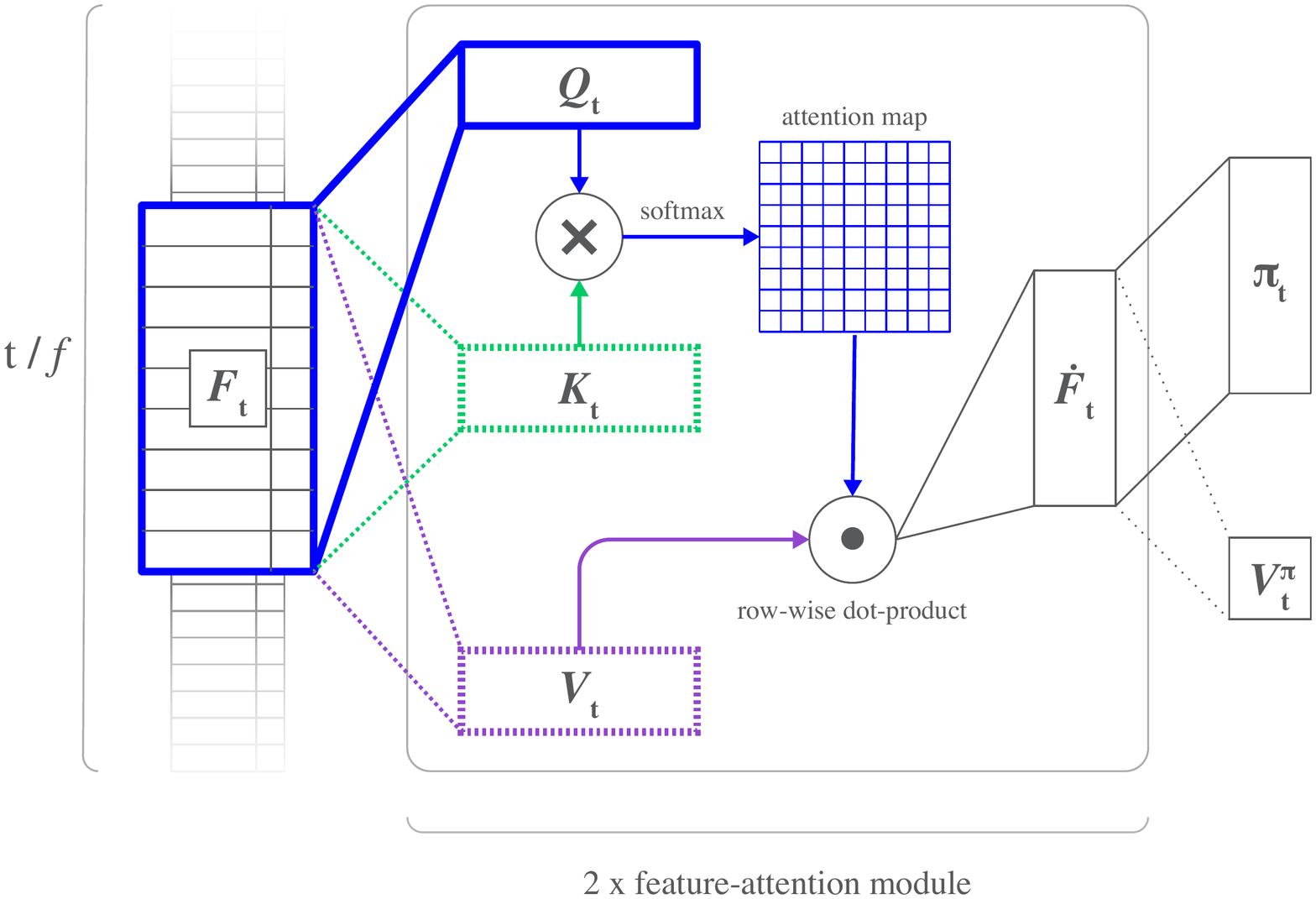}
    \caption{Outline of our self-attention reinforcement learning model. Observations ($ \mathbf{F}_t $) consist of a sequence of $N$ features vectors, across $T$ time-steps. $ \mathbf{Q}_t $ are queries, $ \mathbf{K}_t $ keys, and $ \mathbf{V}_t $ values. Pairwise dot-products of queries and keys generate the attention map, which is then applied to the value representations to produce the weighted compound features. Finally, these compound features inform the policy and value function of the RL agent.}
    \label{fig:model}
\end{figure}

\section{Approach}

Our model (see Figure \ref{fig:model}) implements task-state representations for abstract reasoning using a stack of multi-headed dot-product attention (MHDPA) modules and actor-critic RL. An input observation $ \mathbf{F}_t \in \mathbb{R}^{ [T \times N] \times d_f} $, consisting of a sequence of $N$ feature vectors across $T$ time-steps encoding a specific feature expression, is processed by computing the pairwise interaction between all sequence elements.
In line with the definition of self-attention by \citet{Vaswani2017-ns}, these interactions are differentiable by representing an input sequence element in terms of three distinct embeddings: queries $ \mathbf{Q} \in \mathbb{R}^{ [T \times N] \times d_k} $, keys $ \mathbf{K} \in \mathbb{R}^{ [T \times N] \times d_k} $, and values $ \mathbf{V} \in \mathbb{R}^{ [T \times N] \times d_v} $, parameterized in our case by three distinct sets of trainable weights.
Queries and keys are used to generate attention scores, that is, the mixture weights to represent a sequence element as a weighted sum of all value representations of sequence elements. For a single feature, the attention or relation weights $\mathbf{a}_{i}$ for all other features are expressed by:

\begin{equation}
    \mathbf{a}_{i} = \mathrm{softmax}\Big(\frac{\mathbf{q}_i \mathbf{k}_{1:N}^\mathrm{T}}{\sqrt{d_k}}\Big) \\
\end{equation}

where $\mathbf{q}_i$ is the query representation of feature $i$ and $\mathbf{k}_{1:N}$ are the key representations of all other features. The weighted compound feature is a weighted sum of the whole sequence:

\begin{equation}
    \dot{\mathbf{f}}_i = \mathbf{a}_i\mathbf{v}_{1:N}
\end{equation}

We argue that beyond relating spatially distant feature observations, this process is useful to create compound representations of features across modalities. E.g., relating player position and ball position in Pong (same modality), or compiling ball position and score (different modalities). This inherent quality of dot-product attention has close ties to the task-state-representation framework in neuroscience, which states that attended stimulus dimensions are more readily learned about than others \citep{Leong2017-ps, Niv2015-od, Niv2019-rb}. A high attention score scales the corresponding feature representation up, thus exerting a potentially more significant impact on downstream policy and value functions, which, in turn, motivates a more substantial gradient update for the represented feature.

Just like prior works on self-attention, we define a spatial basis \citep{Vaswani2017-ns, Mott2019-xs}; however, we do not require it to track the sequence position of each element but rather the signature of the attended feature. For simplicity, we use the sequence indices of feature elements in a sliding window of observation. While necessary for effective self-attention, this requirement makes the mechanism extremely resilient towards noise, even after training. Furthermore, this 'signature basis' enables the agent to actively select features during learning, regardless of where they occur in time, as suggested by work in neuroscience \citep{Farashahi2017-ux}.

We next employ the same agent in a partially observable T-maze task to illustrate the capabilities of self-attention to mimic further theories of neuroscience. Specifically, experimental data suggests that the constructs of working memory and attention are closely linked and likely intersect \citep{Chun2011-ae, Chun2011-ji, De_Fockert2001-rw}. Self-attention is demonstrably successful in sequence processing, but, to the best of our knowledge, it has never been applied as a replacement for recurrent models in partially observable, dynamic environments. To test this intuition, we propose a simple task - the T-maze, inspired by rodent experiments (described in detail in the experiments section) \citep{Okada2017-cm}. In the T-maze task, observations $\mathbf{X}
^{T \times H \times W \times c}$ consist of $T$ 2-dimensional image patches of size $H = W = 5$ centered on the agent, with channels representing semantic entities (walls, reward, agent). Therefore, we choose to parameterize the query, key and value functions of this agent with 2d-convolutions.

Both applications of our agent use a stack of two MHDPA modules, as well as two subsequent fully connected layers to compute policy and value function for our actor-critic learner. We trained all our agents, including baselines, using Proximal Policy Optimization (PPO) with both clipped policy and value updates in line with recent commentary on optimal choices for the implementation of PPO \citep{Engstrom2020-nz}.

\section{Experiments}

We test our self-attention agent in a suite of 7 Atari games (see Appendix) with feature-observations and compare learning curves as well as average maximum performance against two baselines. We first run experiments using a simplified version of the model with a single attention head. For the MHDPA, a quick hyperparameter search suggests best performance would be reached with four attention heads per layer.
Both baseline agents are composed of a two-layer MLP, followed by two fully connected layers for both policy- and value- functions. One baseline agent uses convolutional layers, while the other uses fully connected layers, with the number of parameters held approximately constant across our agent and baselines for fair comparison (see Appendix).
Each observation consists of a sequence of $T=4$ time-steps times the number of features provided by the ATARIARI interface \citep{anand2019unsupervised}. We add a number of $n=\{0, 8, 32, 128\}$ noisy features to the observations, and execute 10 full runs, with 4 different random seeds, per condition. Results are reported for $n=\{0, 8, 32\}$ noisy features for the single-headed agent to demonstrate its quick convergence rates at low noise levels, and for $n=\{8, 32, 128\}$ noisy features for the multi-headed model, to show its resilience towards highly noisy observations. All agents are trained for a total number of 1e7 steps in the environment, with learning curve steps averaged over the 10 most recent episodes. Average maximum scores are derived from the 100 best concurrent episodes after 5e6 steps in the environment and averaged over all runs.

For the attention-over-time agent, we compare results against a recurrent agent with two stacked LSTM cells (32 units, two layers). We train each agent for 1e6 steps in the environment, and 10 distinct runs with 4 random seeds. Learning curves are, again, averaged across all runs and and top scores calculated based on the best 100 concurrent episodes.

The attention-over-time agent attempts to solve a T-maze task, inspired by \citet{Okada2017-cm}. The task occurs in a grid-world, with the agent navigating a simple maze, comprising walkable corridors of 9x7 cells, by moving in the four cardinal directions (see Figure \ref{fig:maze} (Left) in the results section). The agent has to follow a specific sequence of moves (a figure-8 over the maze), to generate a collectible reward at one of two preset locations. The field of vision is centered on the agent and extends to a radius of $r=2$ grid cells in each direction. At every time step, the agent needs to compute its next move by processing the past $T=32$ time-steps, which allow the agent to traverse the length of the maze three to four times before an observation leaves the sliding window. Following rodents' experiments \citep{Okada2017-cm}, if the reward location is approached from the wrong side, the state of the maze is reset, forcing the agent to complete a figure-8 without reward, to generate a further collectible reward. The same is true if the agent reaches the top, but elects to go towards the wrong reward location. Both blocked movement actions and noop are possible.

\begin{figure}
    \centering
    \includegraphics[width=0.32\linewidth, trim=5em 3em 5em 3em,]{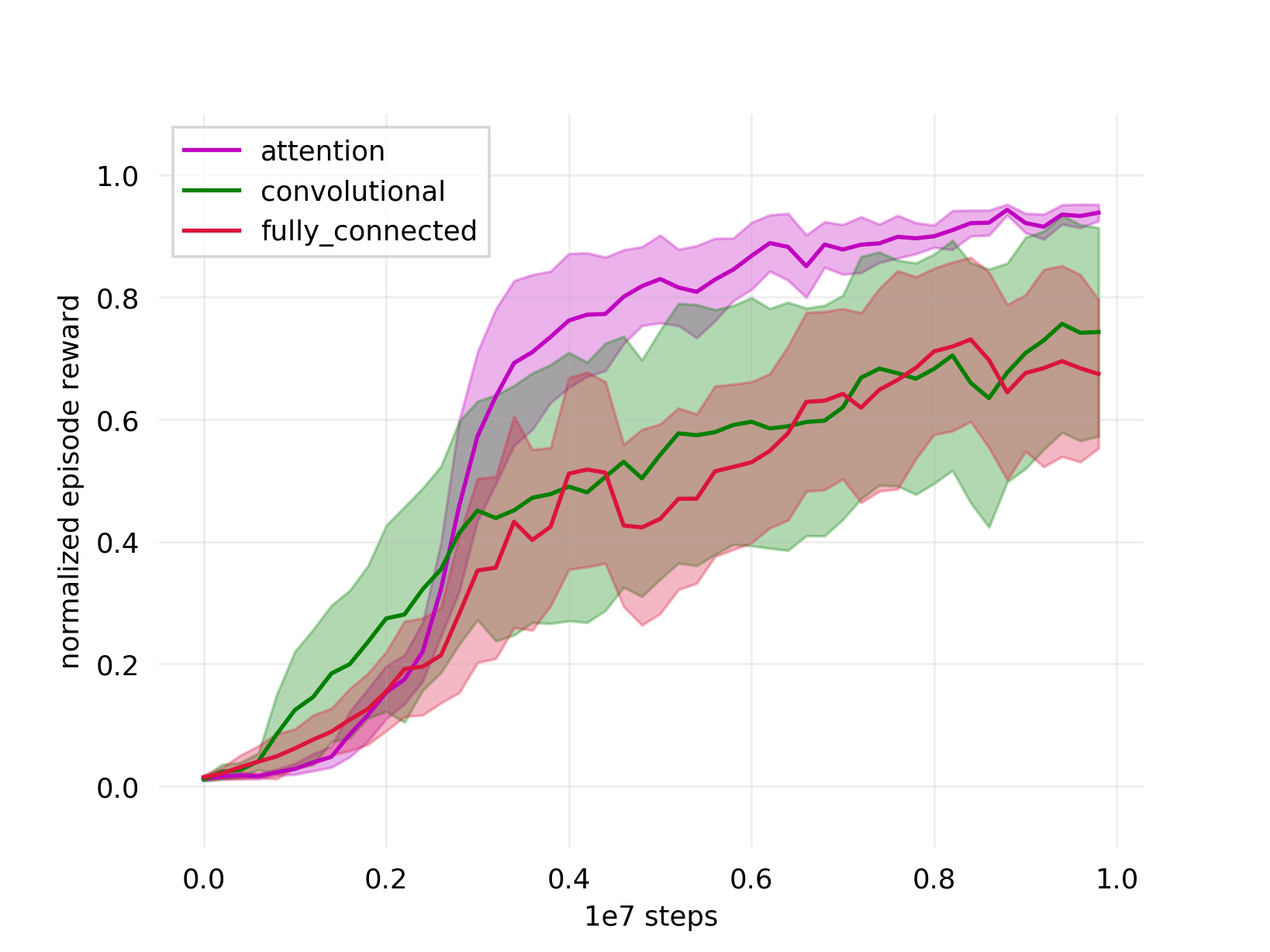}
    \includegraphics[width=0.32\linewidth, trim=5em 3em 5em 3em]{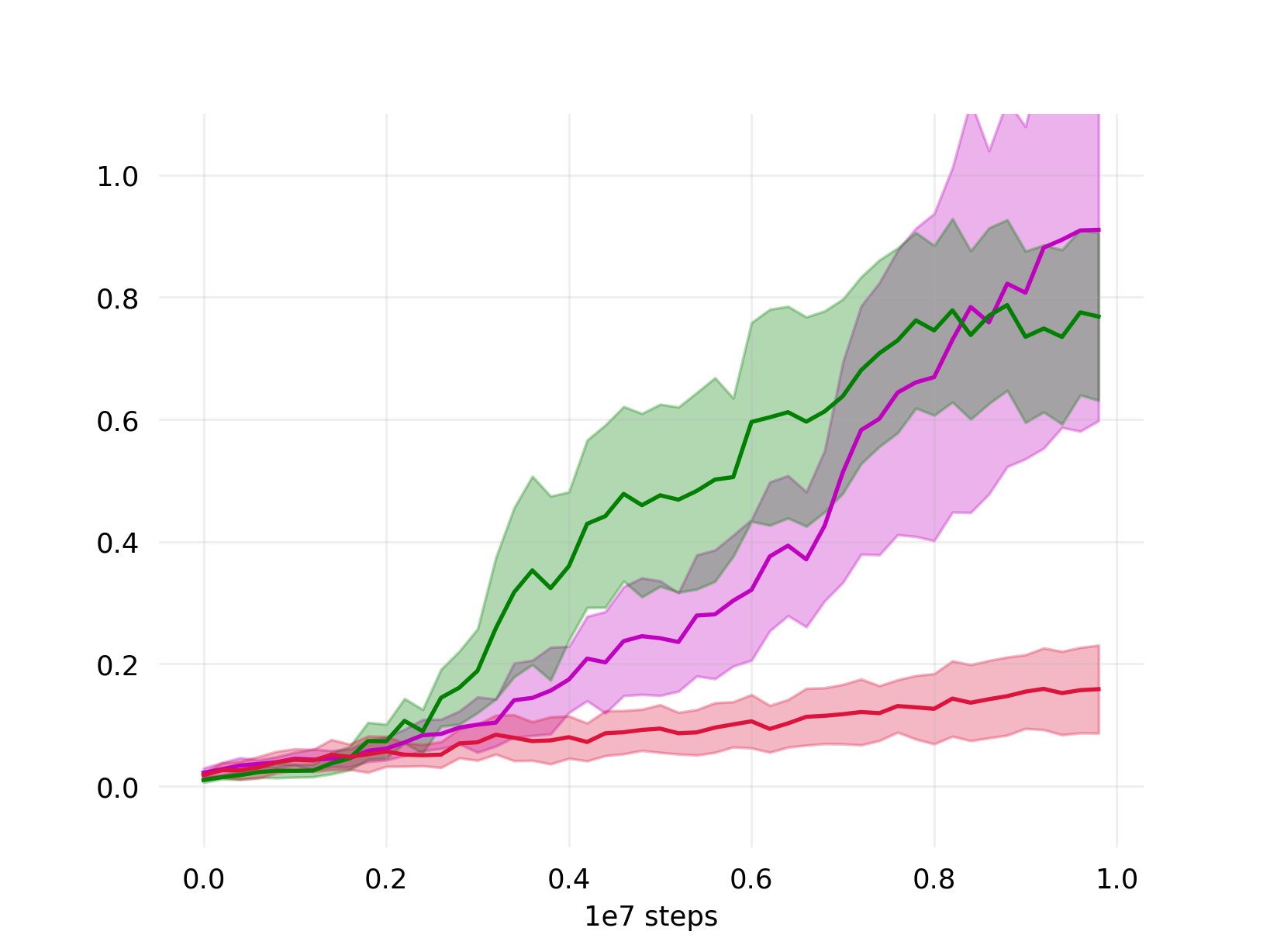}
    \includegraphics[width=0.32\linewidth, trim=5em 3em 5em 3em]{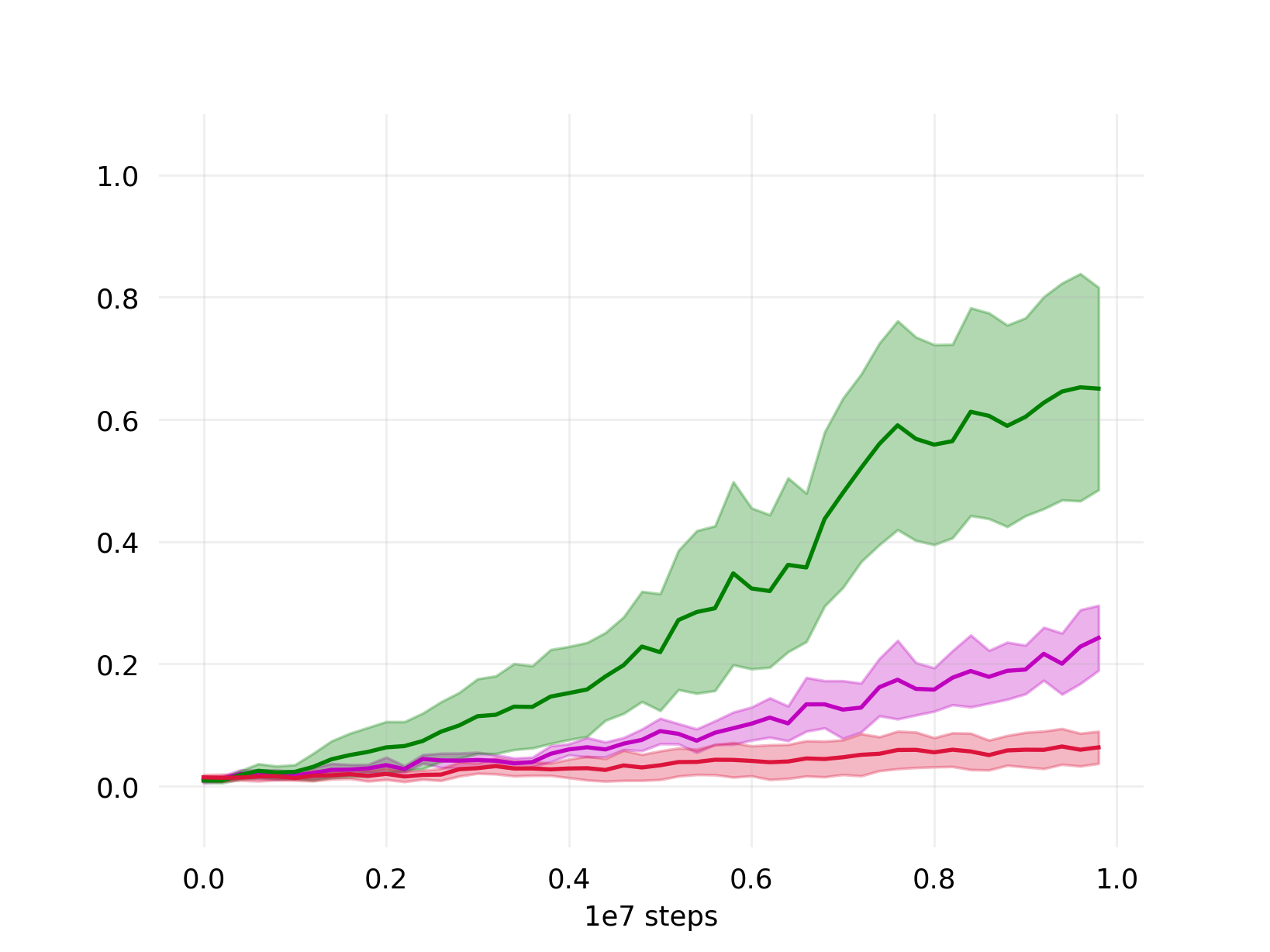}
    \caption{Single-head learning curves in Atari benchmark. Comparison of the single-headed attention agent against both baselines in \{0, 8, 32\} added noisy feature settings. Curves are averaged over 10 distinct runs and represent normalized score across seven Atari games over a total of 1e6 steps per run/environment. The simple attention agent converges quickly in low-noise environments, while the CNN baseline seems more resilient when noise increases.}
    \label{fig:perf0}
\end{figure}

\begin{figure}
    \centering
    \includegraphics[width=0.5\linewidth, trim=5em 2em 5em 5em]{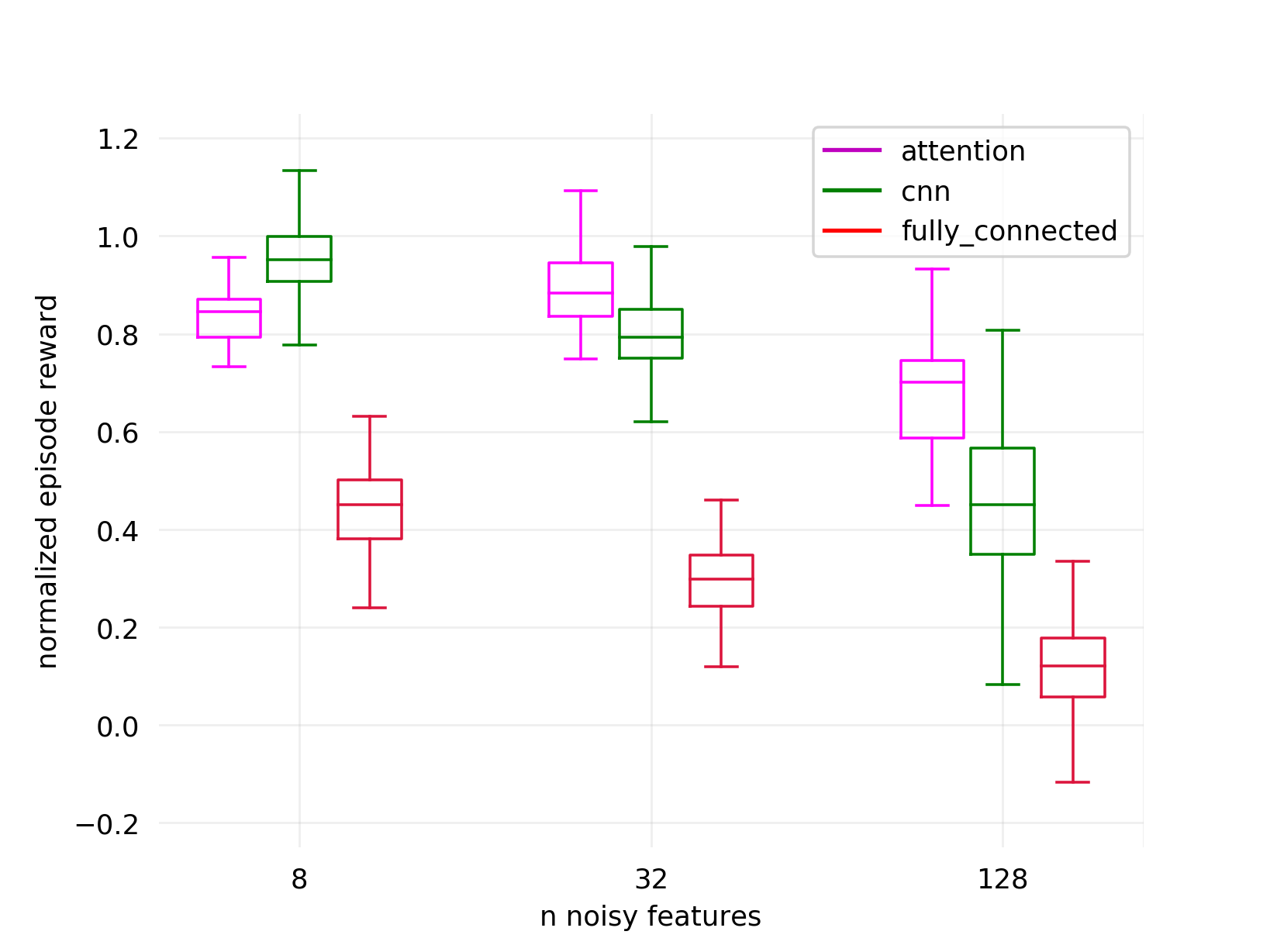}
    \caption{Multi-head maximum performance in Atari benchmark. Comparison of fully trained multi-headed attention agent against both baselines in \{8, 32, 128\} added noisy feature settings. Each trained model was evaluated on 100 episodes per game after training for 1e7 steps. Normalized results were averaged per game and condition. While the multi-headed agent remains relatively slow, it is more effective at maintaining a stable performance in high-noise environments.}
    \label{fig:perf1}
\end{figure}

\begin{figure}
    \centering
    \includegraphics[width=1.\linewidth]{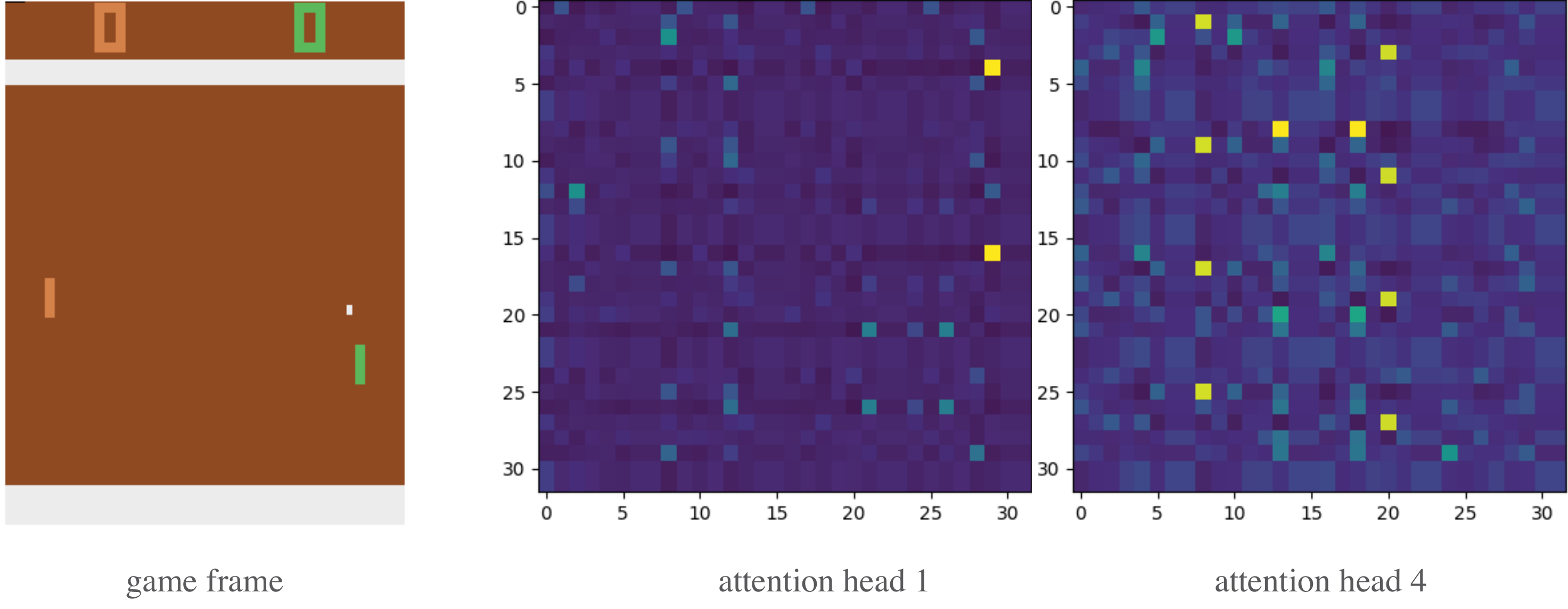}
    \caption{Attention scores in a critical moment. Each cell of the attention map represents a pair of features ($N=8, T=4$) at a certain time-step, with higher index numbers corresponding to later time-steps. Different attention heads track different features and their dynamics. While attention head 1 singularly compiles player-, enemy-, and ball-y-positions to potentially line up a shot, attention head 4 keeps track of player-x-position over time (column index 8), as well as enemy-x-position over time (column index 20).}
    \label{fig:atari_attention_scores}
\end{figure}

\section{Results and Discussion}

Our experiments show promise for the application of self-attention in RL, and for increased cross-seeding between machine learning and neuroscience principles. We detail hereafter four main components of our results.

\subsection{Performance and noisy feature experiments}
First, the single-headed agent displays an increased convergence rate as well as higher peak performance in low-noise environments (see Figure \ref{fig:perf0}), possibly due to its ability to establish non-local connections between elements of the feature sequence. For example, the convolutional agent can only represent local relationships in a single layer and needs at least two layers to relate the first to the last sequence element. By contrast, the multi-headed agent is not as quick but shows remarkable resilience against task-settings with additional noisy features (see Figure \ref{fig:perf1}). We suspected that this was again a result of the ability to model non-local relationships of the self-attention mechanism. To test this theory, we retrained the attention and the convolutional baseline agent on observations, where all task-relevant features were direct sequence neighbors. Indeed, while the self-attention performance remained approximately the same, the performance of the convolutional agent increased (see Appendix).

\subsection{Compiling relevant features}
A primary goal of this feature-based examination of self-attention was to observe feature selection and interaction directly. We find that, generally, attention heads appear to track different feature-constellations in every environment (see Figure \ref{fig:atari_attention_scores}). Across all Atari environments, no attention head is committed to selecting only a single feature; rather, every task-relevant feature is represented as a compound feature. Examples of this include an attention head in Pong tracking ball-x, and enemy-y positions (curiously, this head alone does not yield information on the exact position of the ball, but it may indicate a promising direction to shoot in if the ball is close), in Seaquest, the agent periodically compiles player-y and oxygen-value (indicating that the agent is aware of the oxygen cost of rising to the surface and recharge).

\subsection{Dynamic attentive phases}
Beyond the telling selection and combination of features, we find that the dynamics in attention scores suggest varying attentive phases in the agent. Closely related to the concept of “tripwires”, introduced by Mott et al. \citep{Mott2019-xs}, our agents display attention maps with high entropy until certain conditions are met in the environment. E.g., once the ball in breakout is close enough to the paddle, the attention head representing ball-x, player-x, and ball-y exhibits much higher attention scores for the represented features, and a significant drop in all others. This behavior is certainly enabled by our use of feature-observations, where features can take on different expressions and thus trigger different query, key, and value-representations in self-attention layers. While this interaction may, at first glance, appear to be stimulus-driven and thus an example of bottom-up attention, it exerts much of the theorized properties of top-down attention processes. To use an example from Niv et al. \citep{Niv2019-rb}, if we cross the road, we will pay attention to the position and speed of nearby cars. However, if we search for a taxi, the color becomes a relevant feature as well. Thus, the task ascribes which features and feature expressions become relevant and worthy of attention. Specific limits of feature values in our case induce attentive phases in a more complex task and clearly predict an increase in behavioral expression. The stimulus still triggers a response, but the response is shaped by task-proficiency.

\begin{figure}
    \centering
    \includegraphics[width=0.25\linewidth]{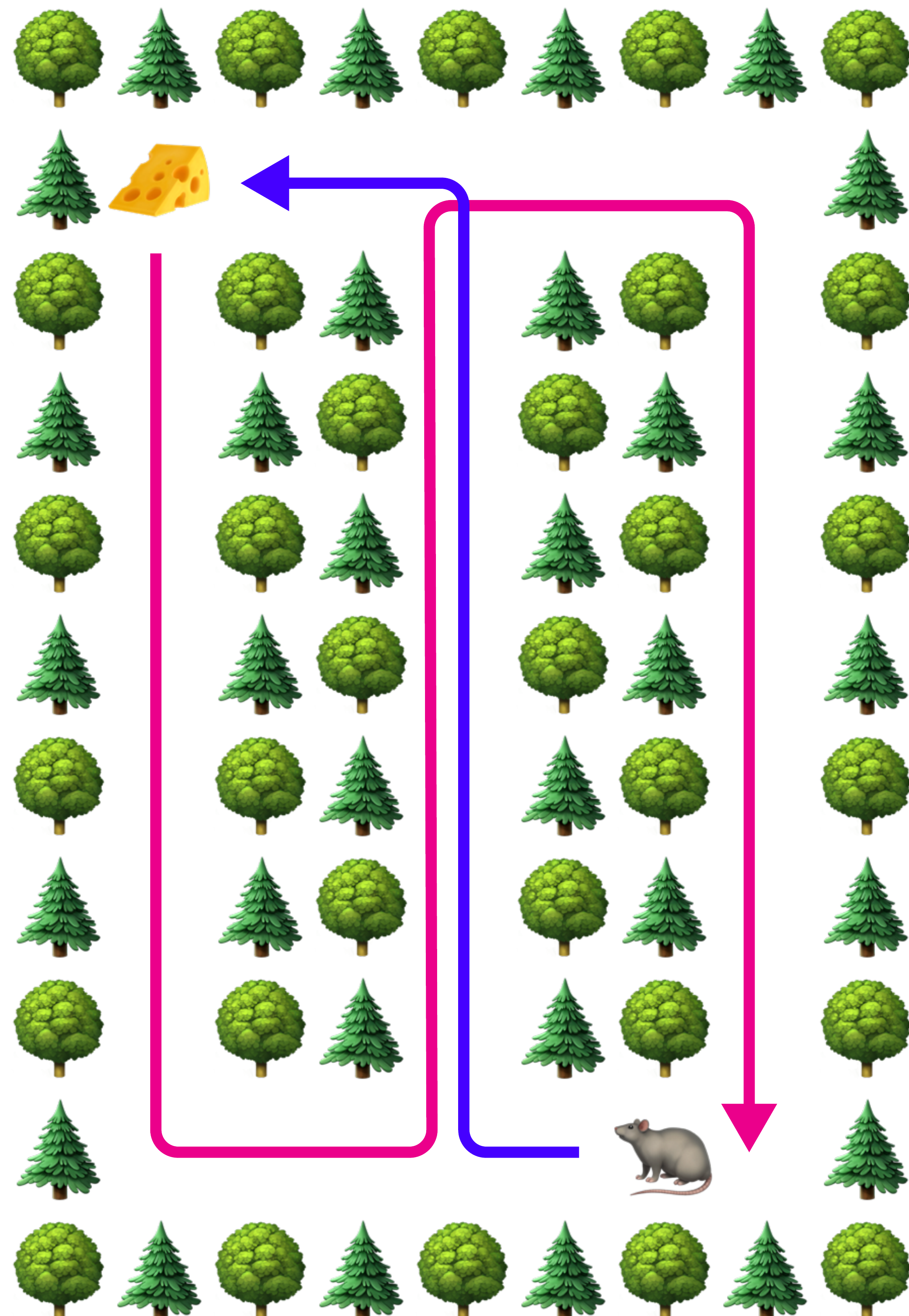}
    \includegraphics[width=0.6\linewidth]{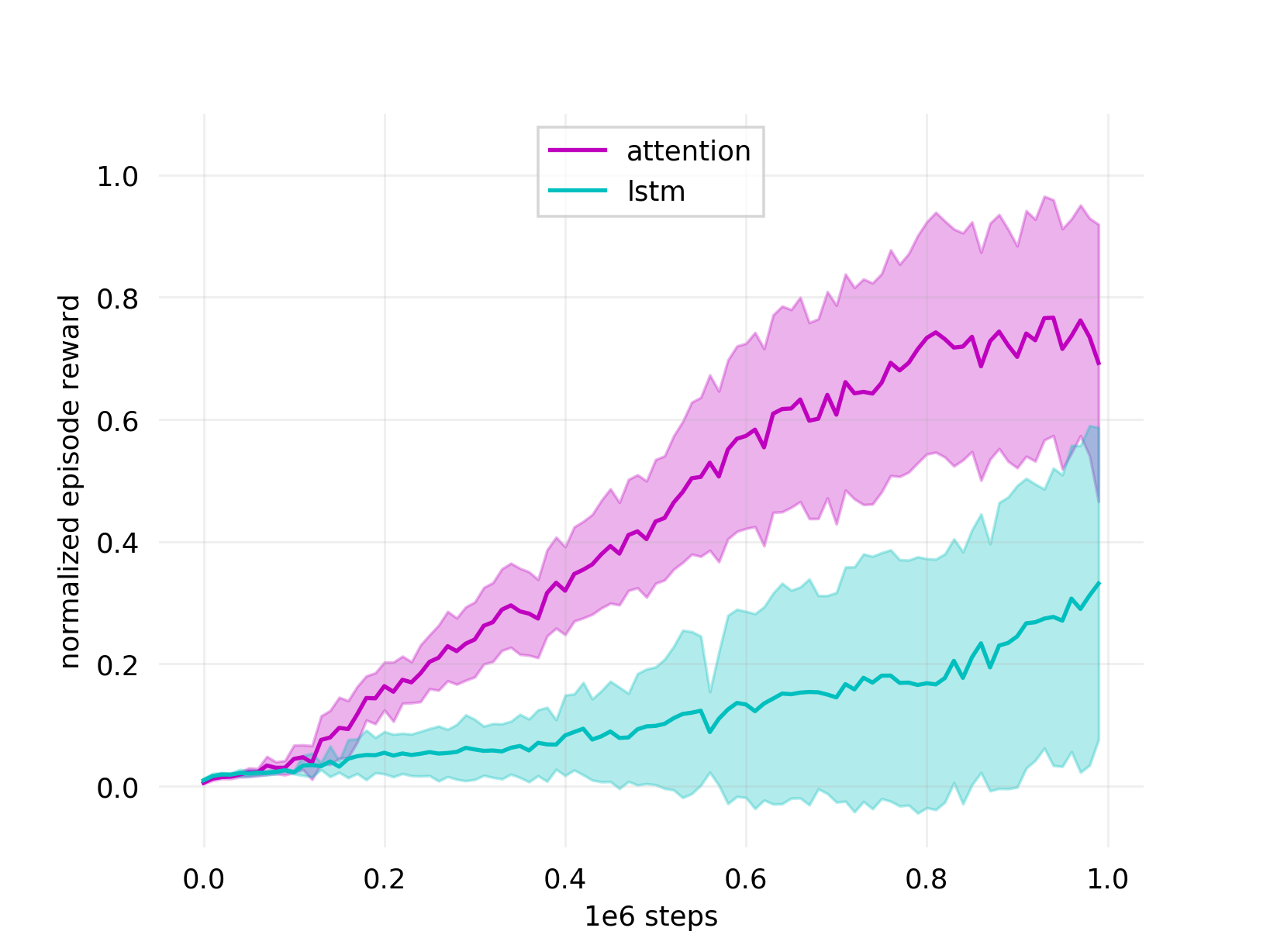}
    \caption{(Left) T-maze task for spatial navigation in rodents. The task is solved by following the optimal path (blue and magenta) and thus triggering reward events at the top left and top right corners of the maze. (Right) Learning curves for the attention-over-time agent against the LSTM baseline.}
    \label{fig:maze}
\end{figure}

\begin{figure}
    \centering
    \includegraphics[width=1.0\linewidth]{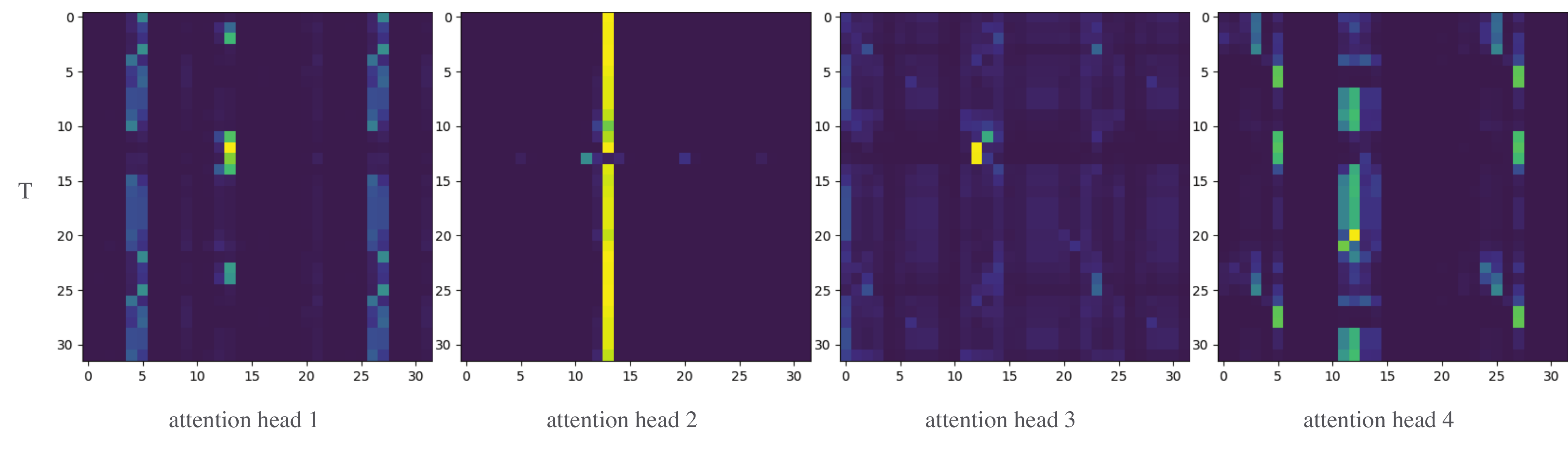}
    \caption{Attention over time in the T-maze. The four attention heads of the agent track past observations to guide orientation. Interestingly, all heads highlight the first observation of the most recent reward object (column index 13). The first and last heads additionally track observations, when the agent was passing the central corridor (column indices (4,5) and (26,27)). Passing of the central corridor and the recent observation of a reward object are attended to at every step of the way (see Appendix).}
    \label{fig:maze_attention_scores}
\end{figure}

\subsection{Attention and Memory}
In our attention-over-time experiment, both agents solve the T-maze task consistently, with the attention agent exhibiting a slight improvement over the baseline's convergence speed (see Figure \ref{fig:maze} (Right)). Interestingly, we do not observe the principle of attentive phases in this agent. Instead, task-relevant information is kept in attentional focus throughout the agent's movement (see Figure \ref{fig:maze_attention_scores}). In the T-maze task, the best indicator for a new reward location is the last reward location. An agent displaying the "tripwire" behavior would likely only focus on the past observation of a reward event, when it needs to decide whether to go left or right. A simple explanation of this behavior could be that the agent is only using the last reward event and its distance in time to locate itself in the maze and choose whether to move up or down an ambiguous vertical corridor. To test this hypothesis, we first trained an attention agent with a single head and found that it was not able to solve the task as effectively as the multi-headed agent (see Appendix). Next, we evaluated a trained agent in a setting where all non-reward event observations were replaced with randomized non-reward event observations. The pre-trained agent was not able to solve the task at all (see Appendix). These results imply that the memory of a past reward event does motivate behavior in this navigation task, but only in conjunction with a more generalized spatial memory it leads to efficient behavior. Lastly, we observed that in cases where the agent took long enough to have the reward-event observation leave the sliding window of observations, the decision to go left or right would be made randomly.

\subsection{Discussion}
The feature-level agent is competitive in solving ATARI games, robust against noise, and, with minor adjustments, extended easily to the temporal domain. Models similar to the one introduced here offer, from a neuroscience or biological viewpoint, a more plausible approach to attention as biological systems continuously generate stimulus representations and should not be drastically affected by learning (representations are present, in spite of task-relevance). This allows for a more efficient and refined sampling strategy, since the system can quickly switch between feature representations depending on task goals. Ultimately though, in the brain, the picture is more complicated: attention and RL interact at several levels of the cortical hierarchy. For example, attention will affect which area of the visual field will have maximal processing of raw sensory information, but also within this sensory stream, which features carry most goal- or reward-related information \citep{Gazzaley2012-qj, Ni5493}.

Attention should enable fast learning, but this is likely only possible when it acts in representational space, rather than immediate sensory space. One limitation of the current work is that features are provided “as is”, i.e., the system does not learn to generate or extract these feature representations. This could be alleviated by using predictive models with strong inductive biases or unsupervised representation learners, such as deep-infomax \citep{anand2019unsupervised} or similar noise-contrastive methods.

Self-attention sequence encoding capabilities translate nicely into RL for decision-making. However, for better or worse, the mechanism lacks a recurrent element, thus discarding full memories without incorporating them into a running context. In cases where memory needs to be more persistent, our model would require a larger horizon. Recent improvements to self-attention promise large-horizon sequence processing with linear complexity, which may be translated to an RL setting.

\section{Conclusion}
We implemented an RL agent that relies heavily on self-attention in feature space, as well as an agent that uses self-attention-over-time to act effectively in partially observable environments. Our results suggest that multi-head-dot-product attention is an effective mechanism to apply reinforcement learning in environments with noisy, feature-based observations, as suggested by contemporary neuroscience literature. Besides, the explicit interpretability of the attention mechanism (in the form of attention maps), combined with its malleability, affords a more accessible perspective on dynamic decision making in artificial agents. We hope that this work may provide useful starting points for future work combining cognitive, psychological constructs with deep reinforcement learning.

\section*{Broader Impact}

Research at the intersection of neuroscience and machine learning or artificial intelligence holds promises. By drawing ideas from neuroscience research, one may find useful mechanisms applicable to ameliorate machine learning models. Although the attention mechanism coupled with reinforcement learning used in this study remains a simple instantiation, this line of work can lead to interesting new concepts and make predictions testable in neuroscience. At the same time, it can be a double-edged sword, as in the end, any advance that makes machine learning models less dependent on sample size, or more resilient to noise, may lead to their use in unethical applications.

\section*{Acknowledgements}
The authors were supported by JST ERATO (Japan, grant number JPMJER1801).

\medskip

\small
\bibliographystyle{plainnat}
\bibliography{main}

\newpage

\section*{Appendix}

\subsubsection{Hyperparameters}

\begin{table}[H]
    \centering
    \begin{tabular}{ l | l }
    \hline
    \bf{Hyperparameter} & \bf{Value} \\ \hline
    \\
    Horizon (T) & 128 \\
    Learning Rate & $2.5 \times 10^{-4}$ \\
    Num. PPO Epochs & 3 \\
    Minibatch Size & $32 \times 8$ \\
    Discount ($\gamma$) & 0.99 \\
    GAE Parameter ($\lambda$) & 0.95 \\
    Number of Actors & 8 \\
    Clipping Parameter & 0.2 \\
    Value Coeff. & 0.5 \\
    Entropy Coeff. & 0.01 \\
    Gradient Norm & 0.5 \\
    \\
    \hline \\
    Number of Attention Heads ($n^h$) & 4 \\
    Query/Key dimensionality ($d_k$) & 32 \\
    Value dimensionality ($d_v$) & 32 \\
    \\
  \end{tabular}
  \caption{Hyperparameters for feature-level Atari experiments.}
\end{table}

\begin{table}[H]
    \centering
    \begin{tabular}{ l | l }
    \hline
    \bf{Hyperparameter} & \bf{Value} \\ \hline
    \\
    Horizon (T) & 128 \\
    Learning Rate & $2.5 \times 10^{-4}$ \\
    Num. PPO Epochs & 3 \\
    Minibatch Size & $32 \times 4$ \\
    Discount ($\gamma$) & 0.99 \\
    GAE Parameter ($\lambda$) & 0.95 \\
    Number of Actors & 4 \\
    Clipping Parameter & 0.2 \\
    Value Coeff. & 1.0 \\
    Entropy Coeff. & 0.01 \\
    Gradient Norm & 0.5 \\
    \\
    \hline \\
    Number of Attention Heads ($n^h$) & 4 \\
    Query/Key dimensionality ($d_k$) & 16 \\
    Value dimensionality ($d_v$) & 32 \\
    \\
  \end{tabular}
  \caption{Hyperparameters for observation-level T-maze experiments.}
\end{table}

\subsubsection{Parameter counts}

\begin{table}[H]
    \centering
    \begin{tabular}{ l | r | r | r | c }
    \hline
    \bf{No. of parameters} & \bf{Inference} & \bf{Policy ($\pi$)} & \bf{Value ($V^{\pi}$)} & \bf{No. of layers} \\ \hline
    \\
    \bf{attention} & $131,712$ & $6,150$ & $1,025$ & 4 \\
    \bf{convolutional} & $125,184$ & $19,974$ & $3,329$ & 4 \\
    \bf{fully-connected} & $146,080$ & $198$ & $32$ & 5\\
    \\
  \end{tabular}
  \caption{Number of learnable parameters by model/sub-network for attention- and baseline-agents in feature-level Atari experiments. Number of layers specifies shared inference layers + separate policy- and value-layers.}
\end{table}

\begin{table}[H]
    \centering
    \begin{tabular}{ l | r | r | r | c }
    \hline
    \bf{No. of parameters} & \bf{Inference} & \bf{Policy ($\pi$)} & \bf{Value ($V^{\pi}$)} & \bf{No. of layers} \\ \hline
    \\
    \bf{attention} & $47,296$ & $5,125$ & $1,025$ & 3 \\
    \bf{lstm} & $42,560$ & $2,885$ & $577$ & 4 \\
    \\
  \end{tabular}
  \caption{Number of learnable parameters by model/sub-network for attention- and baseline-agent in observation-level T-maze experiments. Number of layers specifies shared inference layers + separate policy- and value-layers.}
\end{table}

\subsubsection{(Multi-head attention) Atari benchmark scores}

\begin{figure}[H]
\begin{subfigure}{0.32\textwidth}
\includegraphics[width=1.0\linewidth]{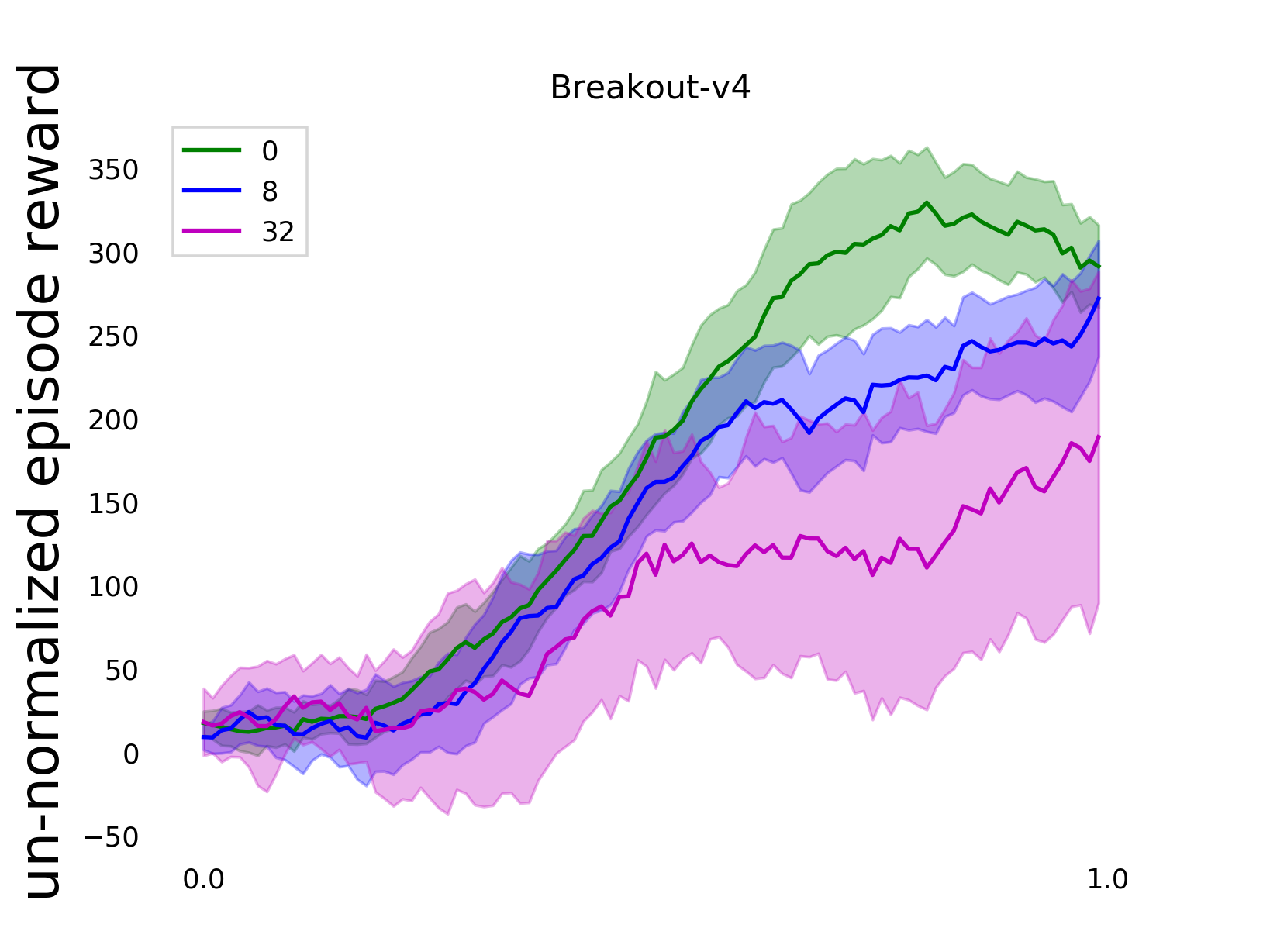}
\end{subfigure}
\begin{subfigure}{0.32\textwidth}
\includegraphics[width=1.0\linewidth]{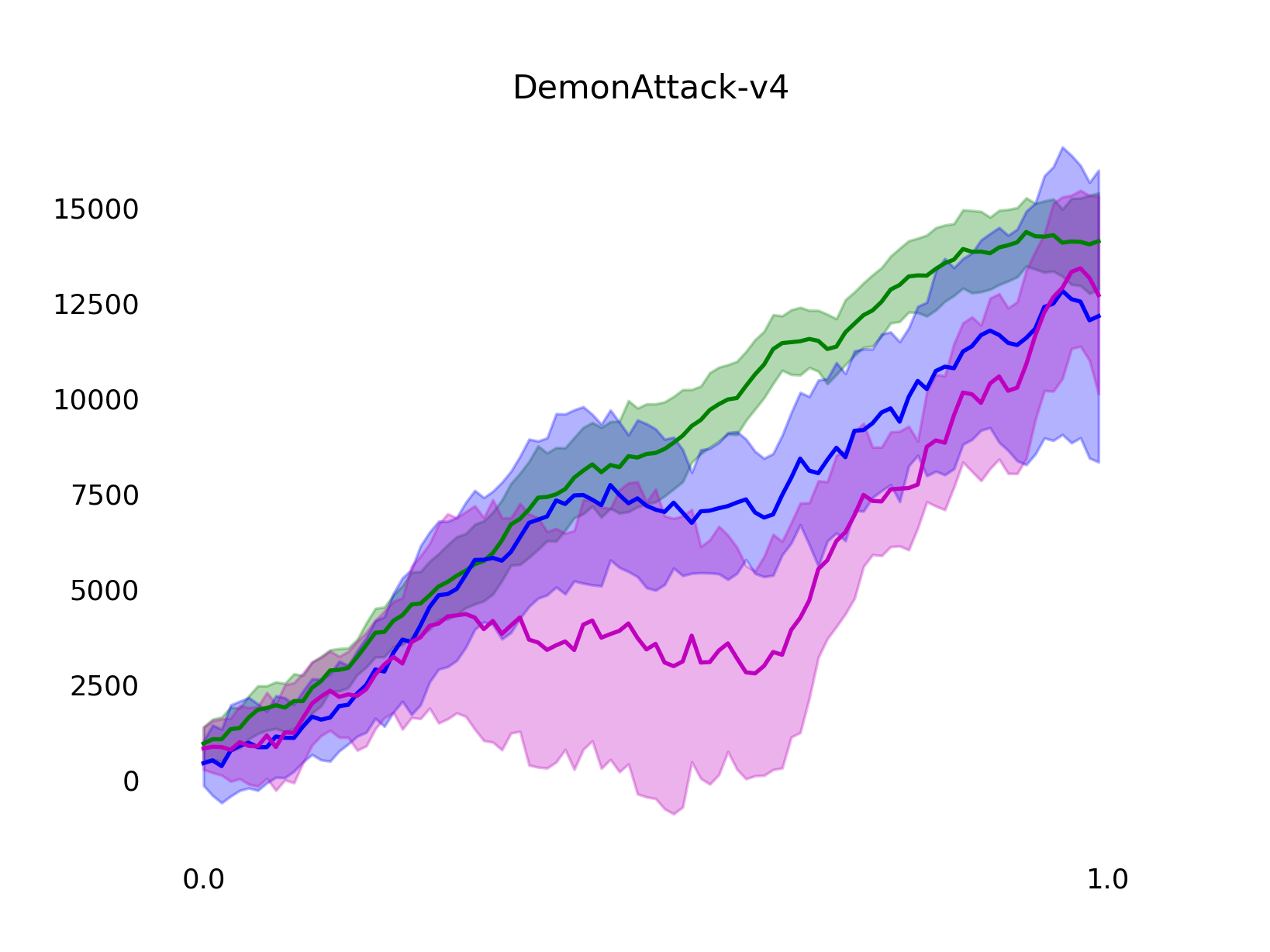}
\end{subfigure}
\begin{subfigure}{0.32\textwidth}
\includegraphics[width=1.0\linewidth]{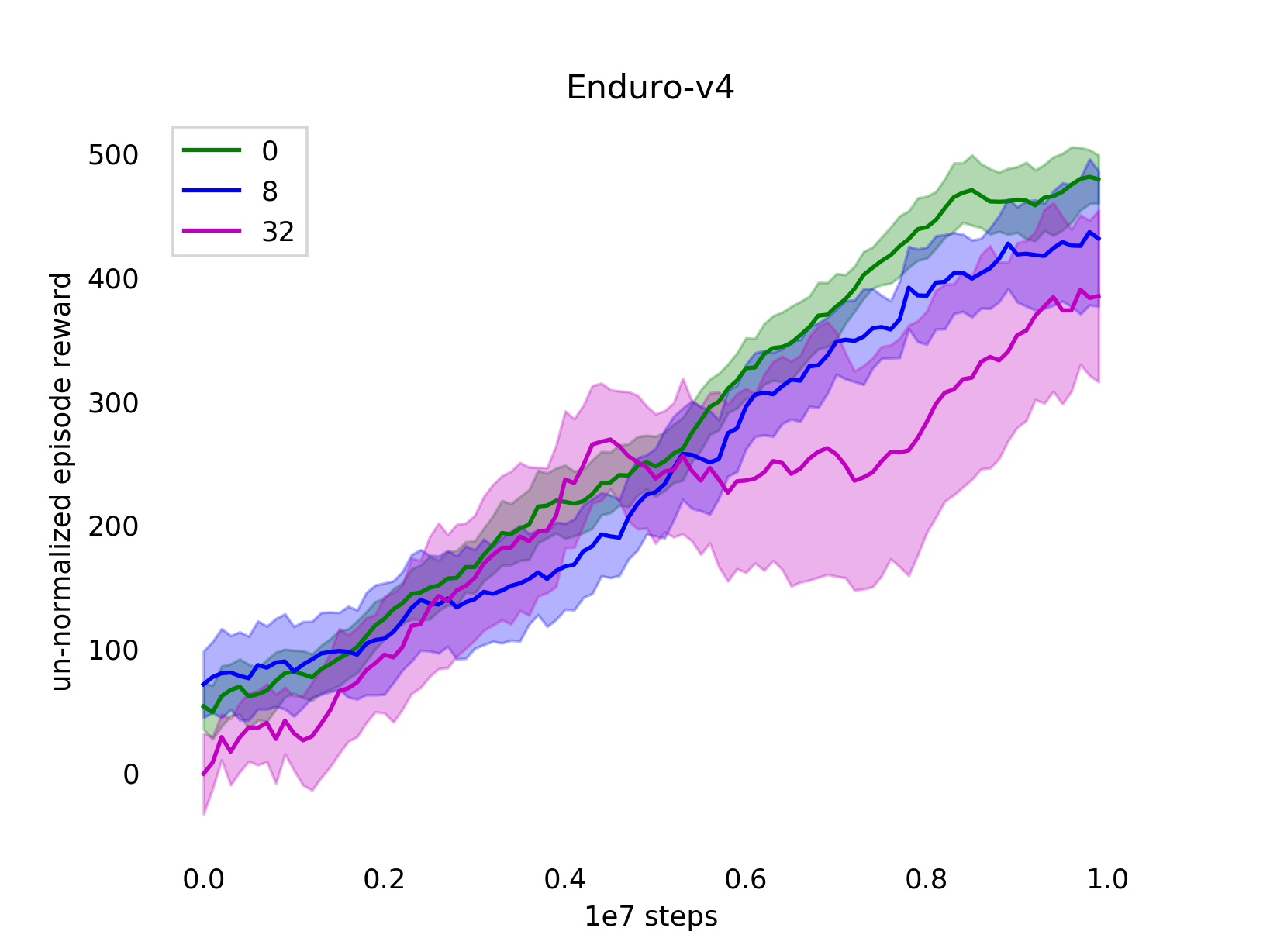}
\end{subfigure}

\begin{subfigure}{0.32\textwidth}
\includegraphics[width=1.0\linewidth]{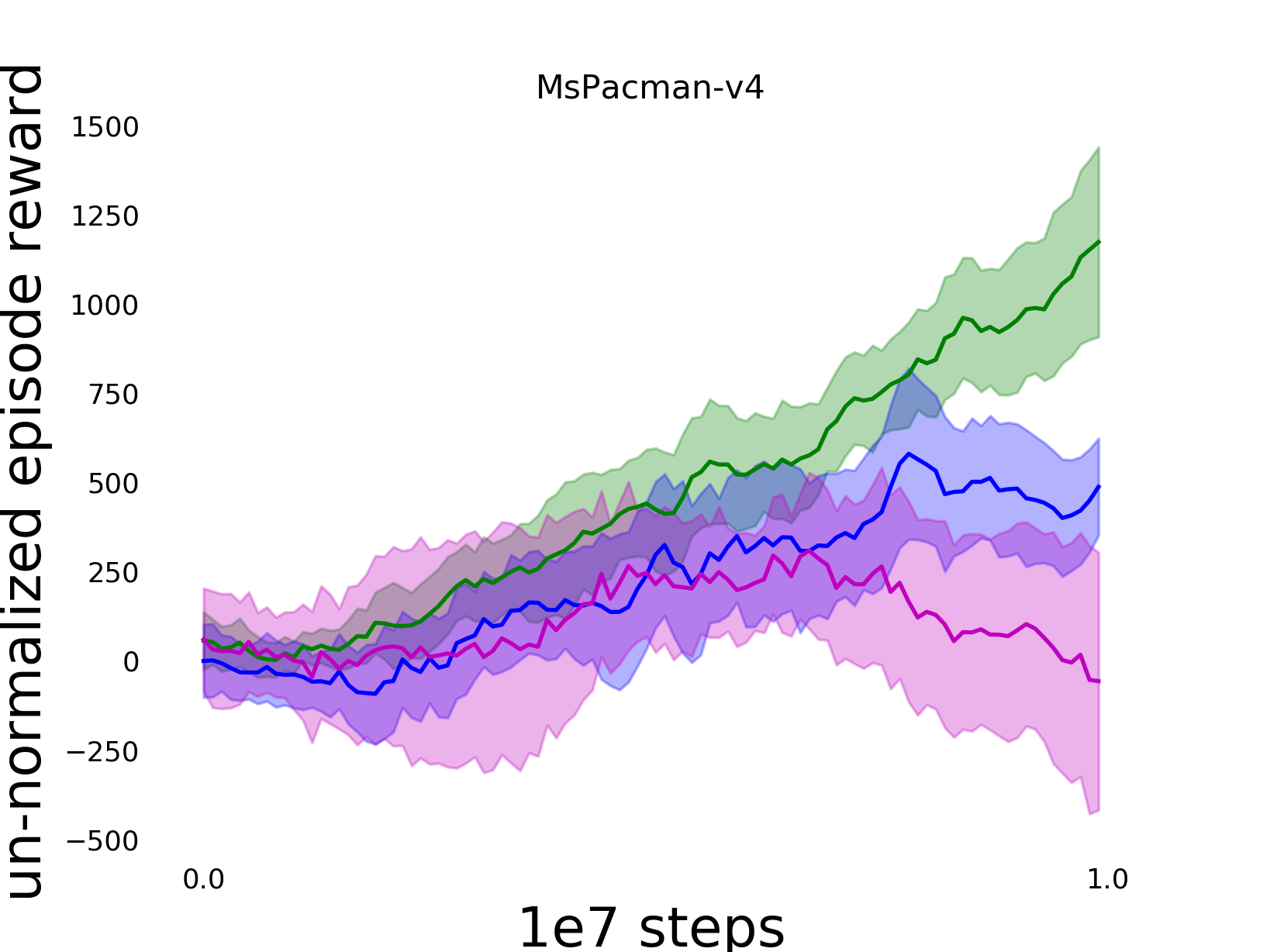}
\end{subfigure}
\begin{subfigure}{0.32\textwidth}
\includegraphics[width=1.0\linewidth]{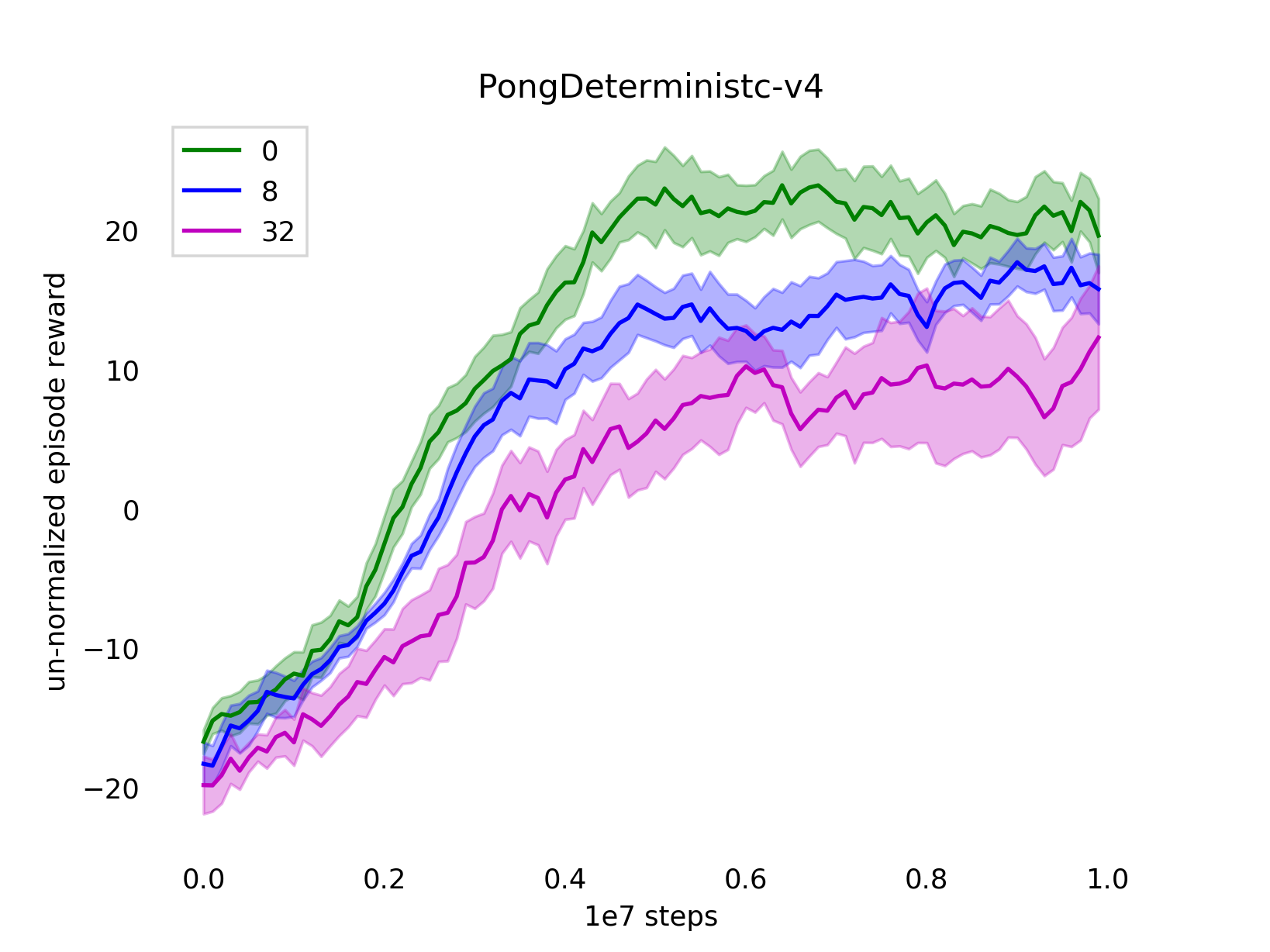}
\end{subfigure}
\begin{subfigure}{0.32\textwidth}
\includegraphics[width=1.0\linewidth]{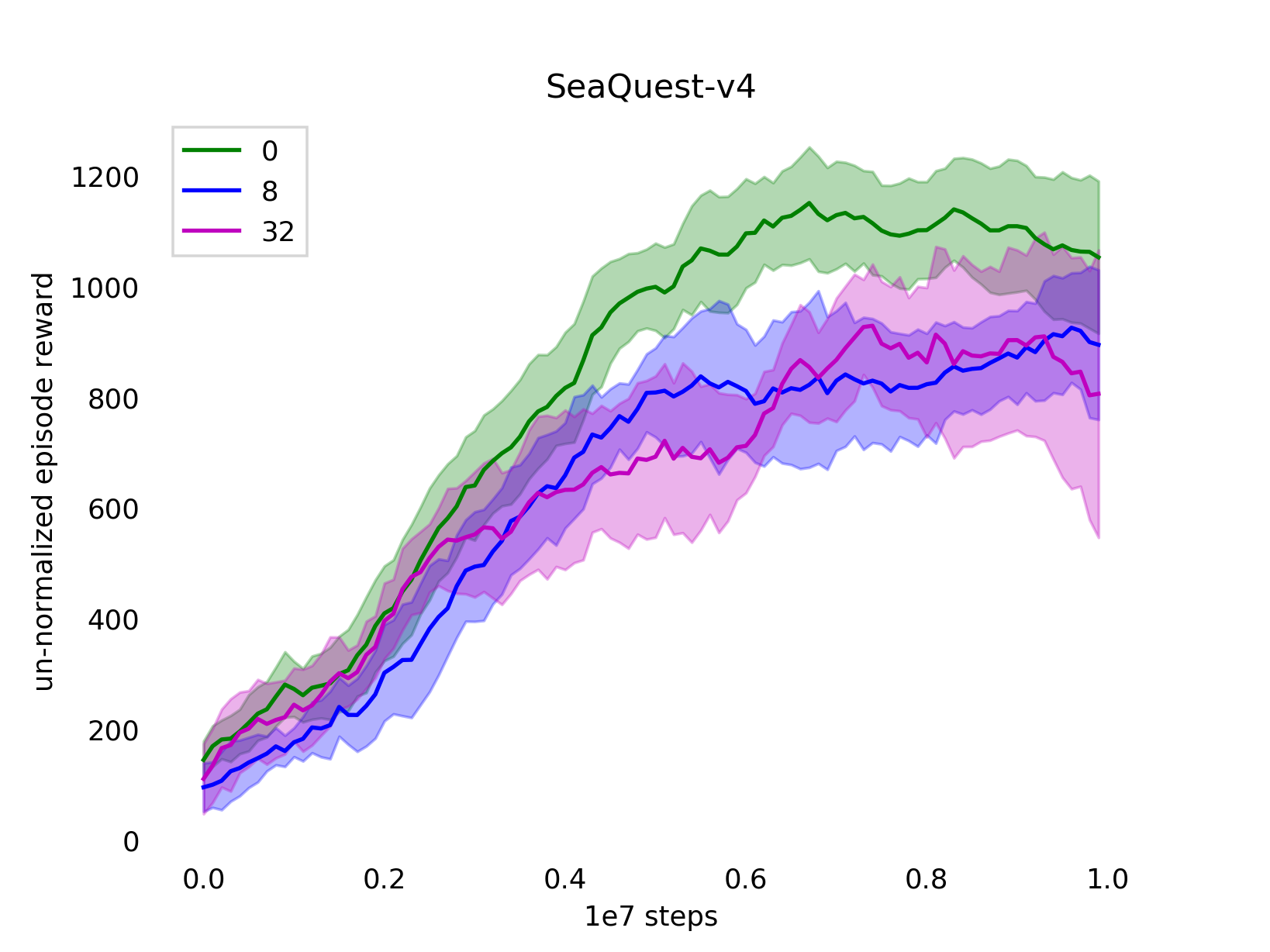}
\end{subfigure}

\begin{subfigure}{0.32\textwidth}
\hfill
\end{subfigure}
\begin{subfigure}{0.32\textwidth}
\centering
\includegraphics[width=1.0\linewidth]{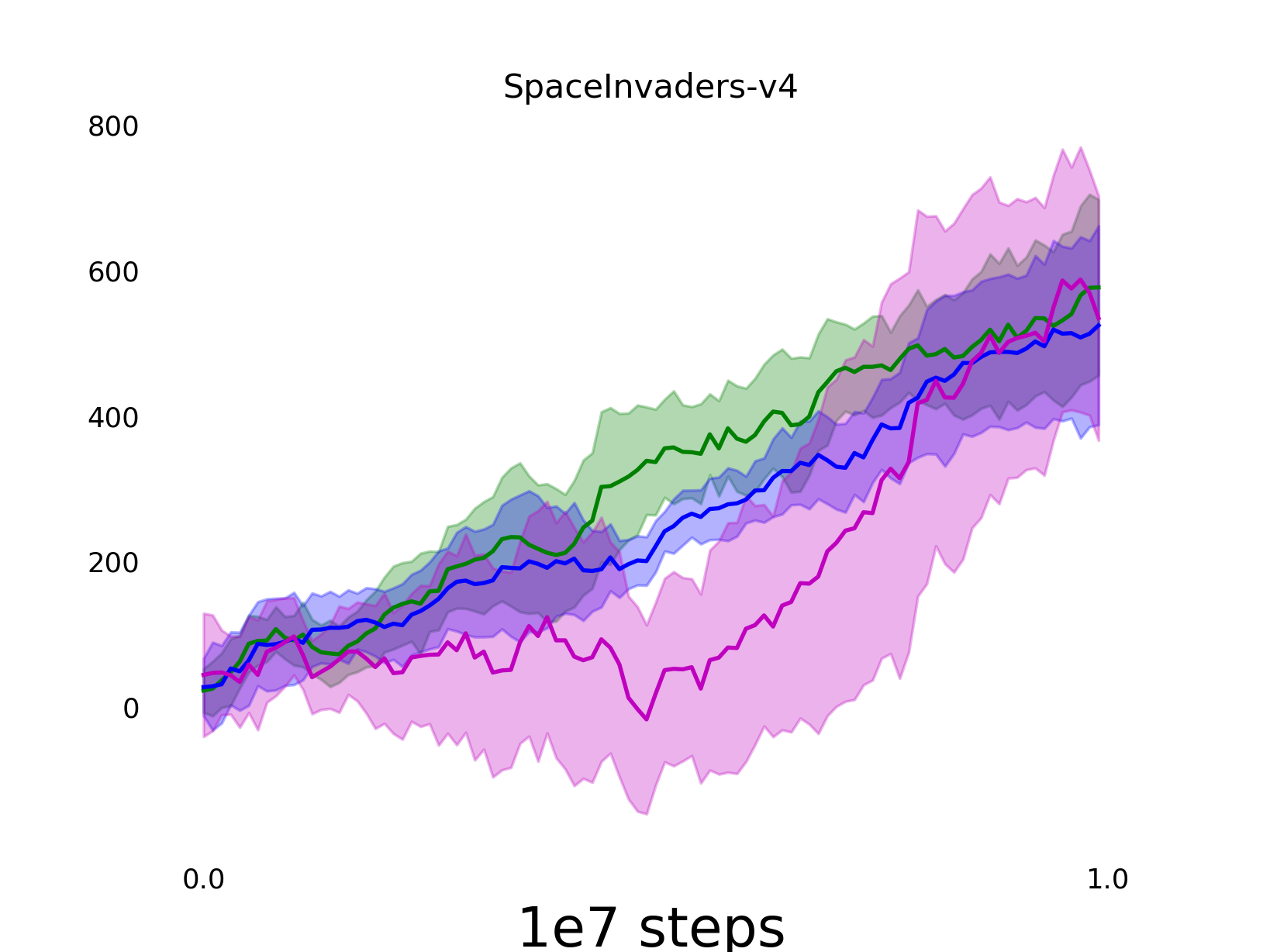}
\end{subfigure}

\caption{Comparison of multi-head attention performances in settings with $n=\{0, 8, 32\}$ noisy features.}
\end{figure}

\subsubsection{(CNN baseline) Performance discrepancy for ordered feature sequence}

\begin{figure}[h]
    \centering
    \includegraphics[width=0.8\linewidth, trim=0em 1em 3em 4em, clip]{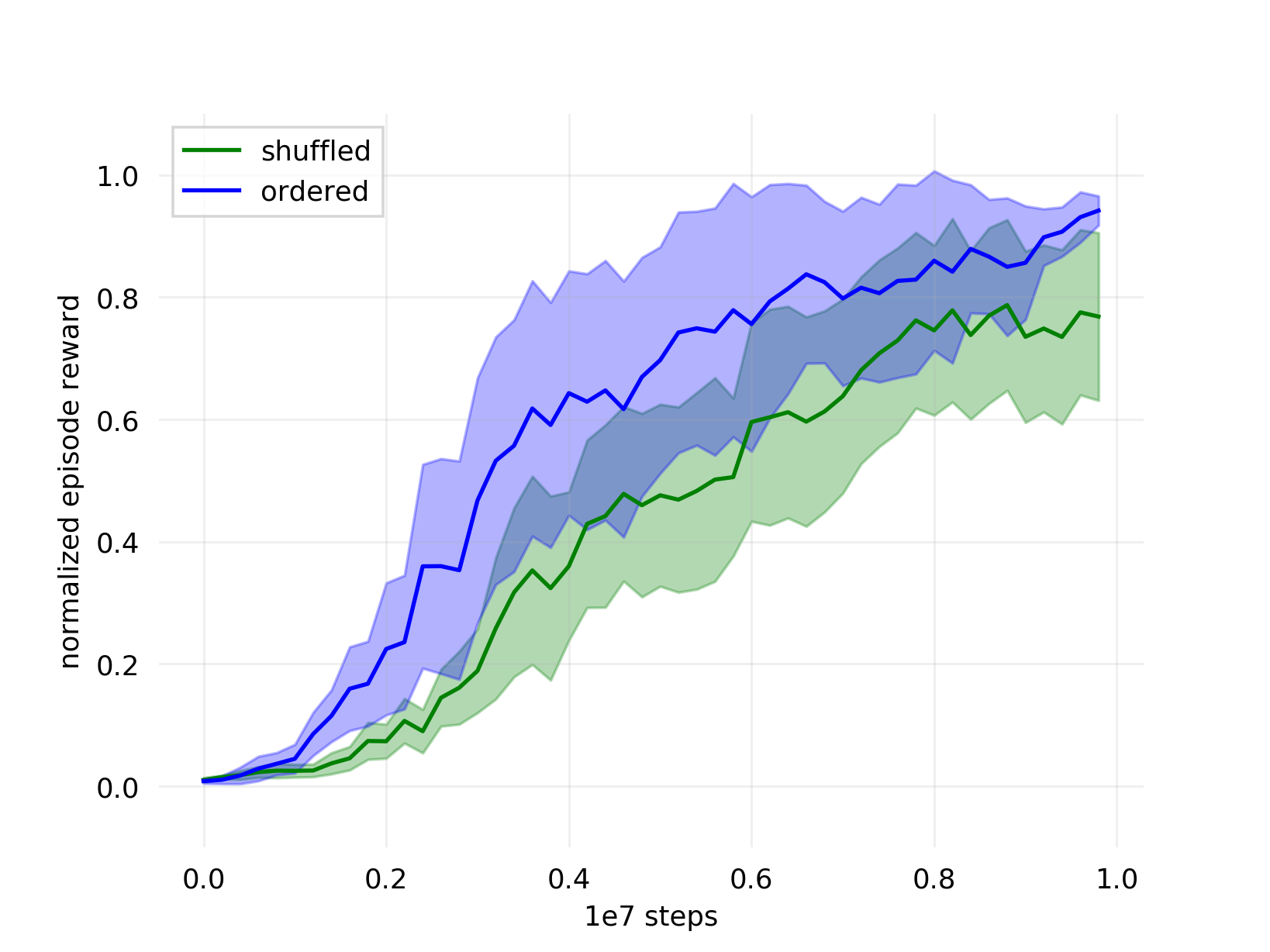}
    \caption{In our experiments, feature sequences were shuffled with a key generated at the beginning of the run. We expected that close neighbourhoods between task-relevant features would improve learning performance in the CNN baseline. Results from 10 experiments with un-shuffled features in a noisy setting $n=\{8\}$ of the game 'Pong' support this hypothesis.}
    \label{fig:shuffled-atari}
\end{figure}

\subsubsection{(Multi-head attention over time) Multi- vs. single-head attention in T-maze task}

\begin{figure}[H]
    \centering
    \includegraphics[width=0.8\linewidth, trim=0em 1em 3em 4em, clip]{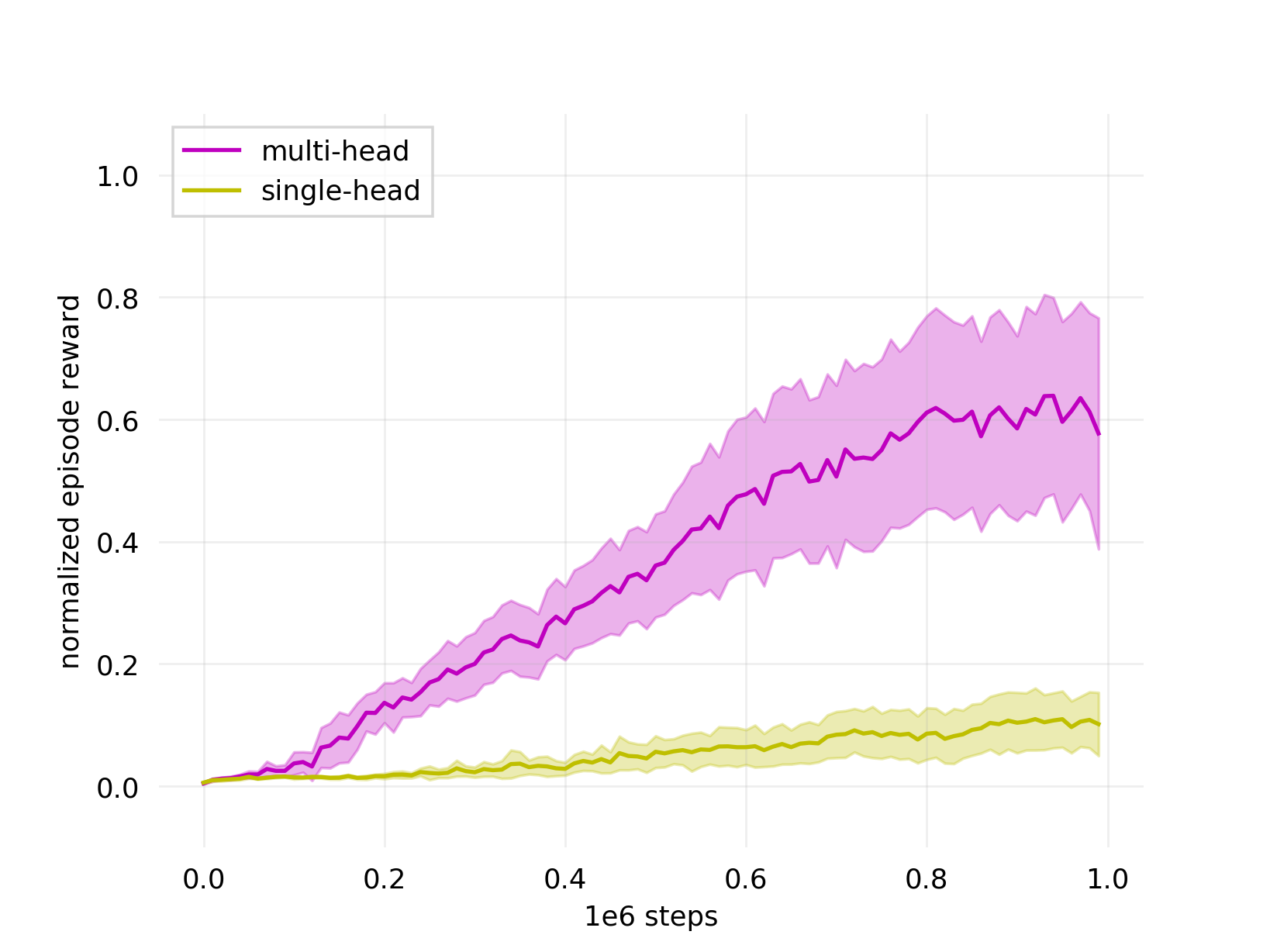}
    \caption{We ran a single-head agent in the T-maze task, to explore whether remembering only the most recent goal location would allow the agent to solve the environment. Unlike in our early Atari experiments, the single-headed agent earns only a small fraction of the possible score.}
    \label{fig:multi-single-maze}
\end{figure}

\subsubsection{Full T-maze sequence}

\begin{figure}[H]
    \includegraphics[width=1.0\linewidth, trim=12em 1em 12em 1em, clip]{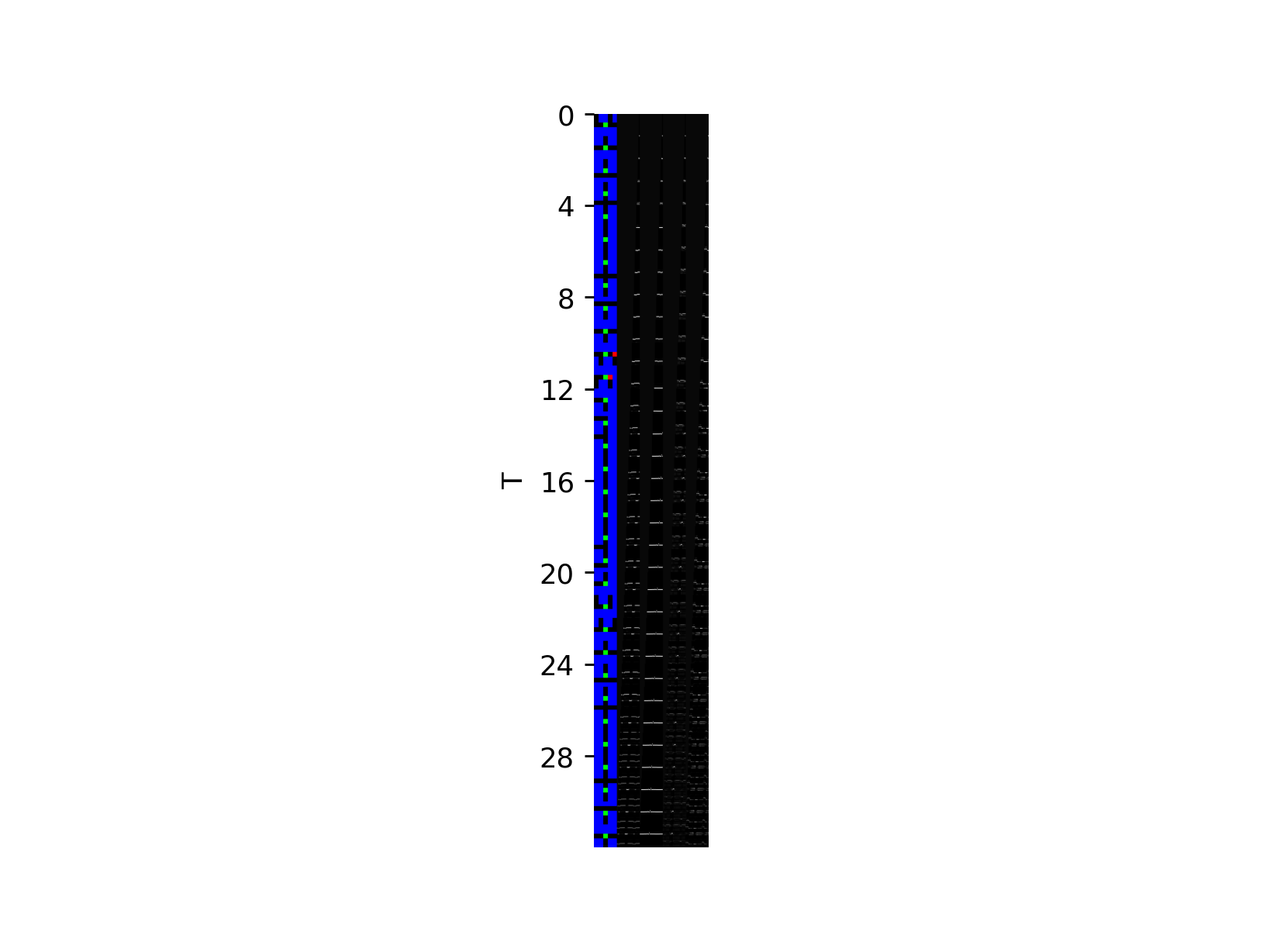}
    \caption{Sequence input in the maze task on the left, 4 concatenated attention heads, corresponding to sequence step, on the right. Red pixels indicate a visible reward object. The second head from the left tracks reward observations.}
    \label{fig:multi-single-maze}
\end{figure}

\subsubsection{Randomized non-reward observations}

\begin{figure}[H]
    \includegraphics[width=1.0\linewidth, trim=0em 0em 3em 4em, clip]{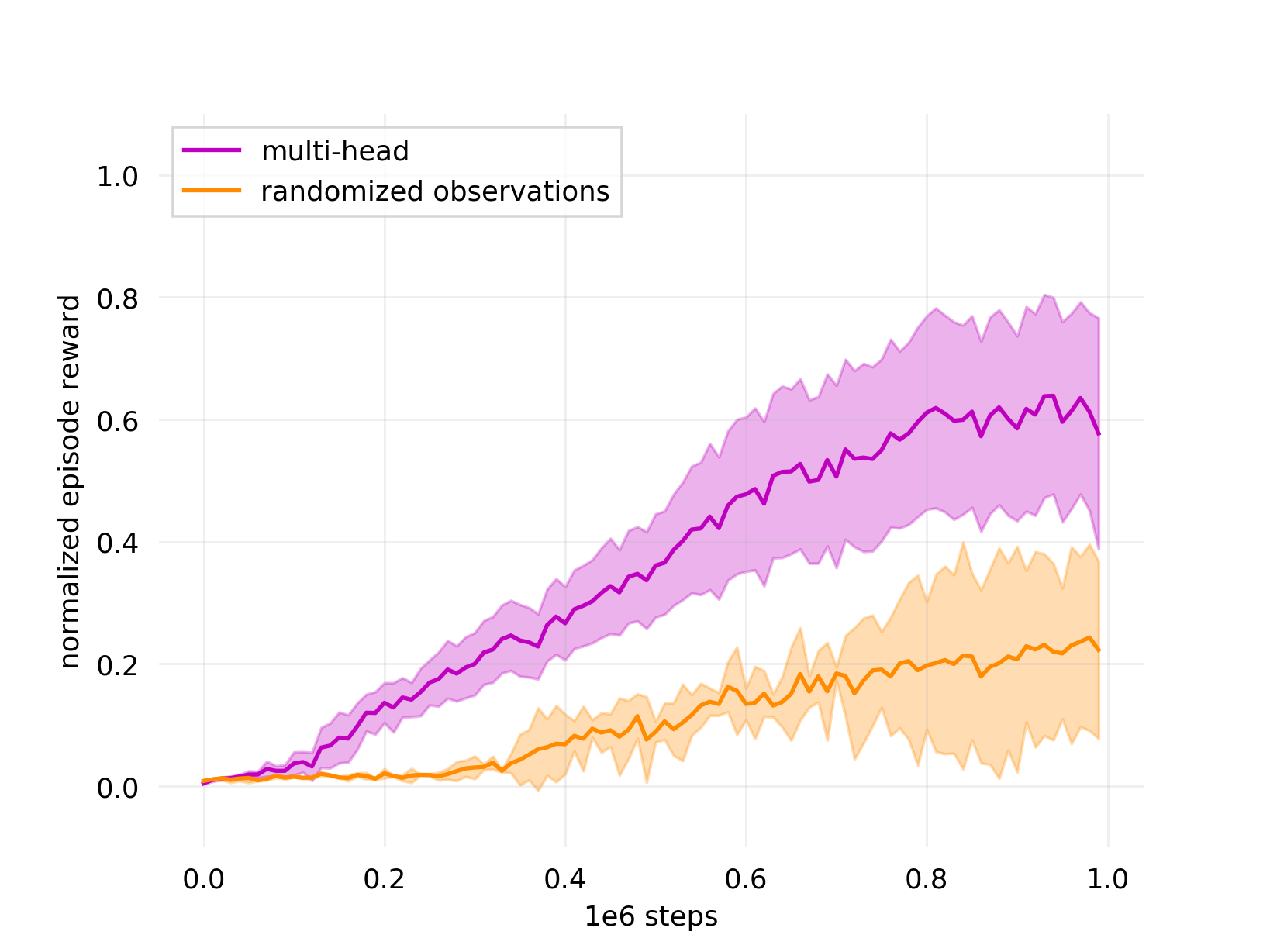}
    \caption{Pre-trained agents rely heavily on non-reward events in memory. However, an agent with randomized non-reward observations is able to display effective learning behavior, given sufficient time - possibly by relying on the timespan between experienced reward events.}
    \label{fig:multi-single-maze}
\end{figure}

\end{document}